\documentclass[runningheads]{llncs}
\usepackage{graphicx}

\usepackage{tikz}
\usepackage{comment}
\usepackage{amsmath,amssymb} 
\usepackage{color}

\usepackage[accsupp]{axessibility}  

\newcommand{\diyi}[1]{}
\newcommand{\james}[1]{}
\newcommand{\patsorn}[1]{}
\newcommand{\witta}[1]{}

\newcommand{\attend}[1]{\textcolor{red}{[#1]}}
\newcommand{\ourmethod}{TASK-former}

\newcommand{\projecturl}{\url{https://janesjanes.github.io/tsbir/}}

\newcommand{\titledpar}[1]{\vspace{7pt}\par\noindent\textbf{#1}}

\usepackage{booktabs}
\usepackage{caption}
\usepackage[capitalize]{cleveref}
\usepackage{multirow}
\usepackage{xcolor}
\usepackage{url}

\crefname{section}{Sec.}{Secs.}
\Crefname{section}{Section}{Sections}
\Crefname{table}{Table}{Tables}
\crefname{table}{Tab.}{Tabs.}

\begin{document}
\pagestyle{headings}
\mainmatter
\def\ECCVSubNumber{7388}  

\title{A Sketch Is Worth a Thousand Words:\\Image Retrieval with Text and Sketch} 

\titlerunning{Image Retrieval with Text and Sketch}
%
\author{Patsorn Sangkloy\inst{1,3}
\and
Wittawat Jitkrittum\inst{2}
\and
Diyi Yang\inst{1}
\and
James Hays \inst{1}
}
\authorrunning{P. Sangkloy et al.}
%
\institute{Georgia Institute of Technology, Atlanta GA 30332, USA
\and
Google Research, New York NY 10011, USA\\
 \and
Phranakhon Rajabhat University, Bang Khen Bangkok 10220, Thailand\\[2mm]
\email{patsorn.s@pnru.ac.th}, \quad\email{wittawat@google.com}, \\
\email{diyi.yang@cc.gatech.edu}, \quad\email{hays@gatech.edu}
}

\maketitle
\begin{abstract}
   We address the problem of retrieving in-the-wild images with \emph{both a sketch and a text query}. We present TASK-former (Text And SKetch transformer), an end-to-end trainable model for image retrieval using a text description and a sketch as input.
   We argue that both input modalities complement each other in a manner that cannot be achieved easily by either one alone. \ourmethod{} follows the late-fusion dual-encoder approach, similar to CLIP \cite{clip-radford21a}, which allows efficient and scalable retrieval since the retrieval set can be indexed independently of the queries. We empirically demonstrate that using an input sketch (even a poorly drawn one) in addition to text considerably increases retrieval recall compared to traditional text-based image retrieval. To evaluate our approach, we  collect  5,000  hand-drawn sketches for images in the test set of the COCO dataset. The collected sketches are available a \projecturl{}.
\end{abstract}

\begin{center}
\centering
\includegraphics[width=\textwidth]{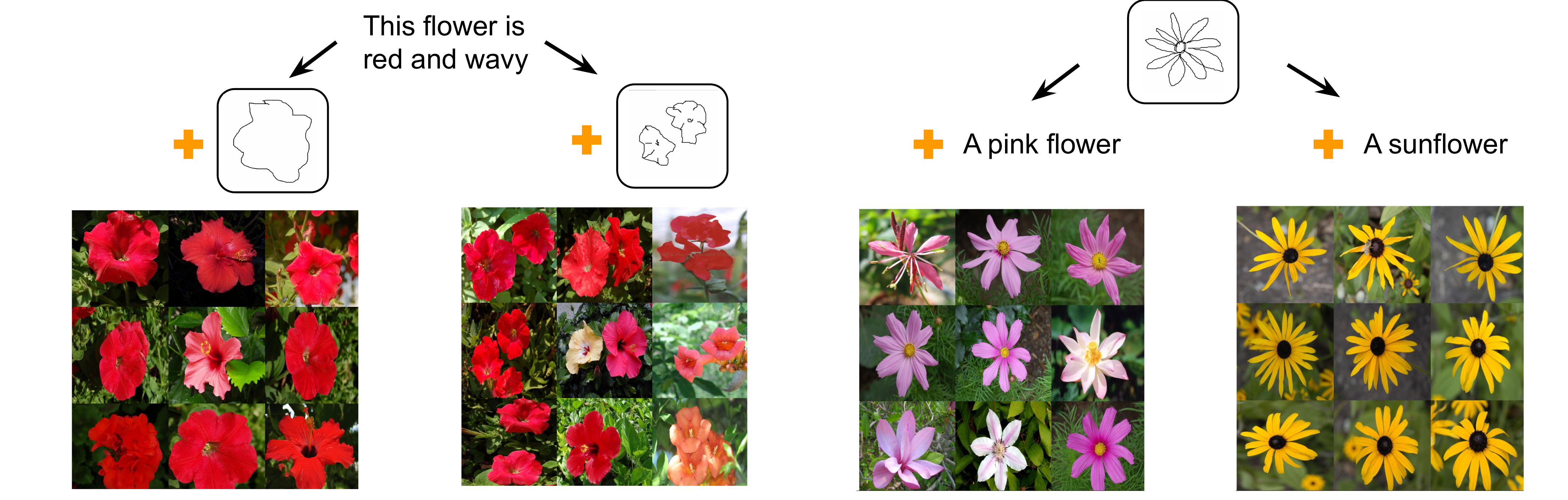}
\captionof{figure}{We present \ourmethod{}, a Text And SKetch transformer based method
for image retrieval. We demonstrate that the presence of a sketch input, even a poorly drawn one, helps narrow down the set of retrieved images to ones that match the joint description provided by the sketch and the text query.
Retrieval results are from our model trained on Flower102 dataset \cite{nilsback2008automated}
}.
\label{fig:teaser}
\end{center}

\section{Introduction}
\label{sec:intro}


Cross-modal retrieval \cite{WanYanXuHan2017} is a retrieval problem where the
query and the set of retrieved objects take different forms. A representative
problem of this class is Text-Based Image Retrieval (TBIR), where the goal is to
retrieve relevant images from an input text query. TBIR has been studied
extensively, with recent interest focused on transformer-based models
\cite{radford2019language}.
A core component of a cross-modal retrieval model is its scoring function that
assesses the similarity between a text description and an image.
Given a text description, retrieving relevant images then amounts to
finding the top $k$ images that achieve the highest scores.


While a text description is suitable for describing qualitative attributes of an
object (e.g., object color, object shape) in TBIR, it can be cumbersome when there is a
need to describe multiple objects or a complex shape. Consider the query ``This
flower is red and wavy’’ in \Cref{fig:teaser} as an example. This query alone
can match red flowers of a wide range of shapes. Further, with multiple objects,
it becomes necessary to describe their relative positions, making the task too
cumbersome to be practical. These limitations naturally led to a related thread of
research on Sketch-Based Image Retrieval (SBIR), where the goal is to retrieve
images from an input sketch
\cite{sketchy2016,pandey2020stacked,yu2016sketch,dutta2019semantically,Pang_2019_CVPR,ZhaZhaFenZha2020,SaiBhuYanXia2021}.
Compared to text, specifying object positions with a hand-drawn sketch is relatively easy.

Recent works on SBIR tend
to focus on a specific set of retrievable images. For instance,
\cite{sketchy2016} considers object based retrieval: each image has
only one object centered in the image. Compared to \cite{sketchy2016},
\cite{yu2016sketch} allows a more free-form sketch input, but from a single
category e.g., shoes. While suitable for an e-commerce product search, searching
from an arbitrarily drawn sketch, also known as  \emph{in-the-wild SBIR}, was
not studied in \cite{yu2016sketch}. In-the-wild image retrieval is the problem
we tackle in this work.

In-the wild SBIR presents two challenges.
Firstly, there is semantic ambiguity in a sketch drawn by a non-artist user. To a
non-artist, it takes effort to draw a sketch which sufficiently accurately
represents the desired image to retrieve.  While adding details to the
sketch would naturally narrow down the candidate image set, to a non-artist
user, the extra effort required to do so may outweigh the intended convenience of SBIR.
Ideally the retrieval system should be able to extract relevant information from
a poorly drawn sketch.
Secondly, the
few publicly available SBIR datasets contain only either images of single
objects \cite{sketchy2016,yu2016sketch}, or images describing single concepts
\cite{flickr15k}. These images are different from target images in in-the-wild
image search, which may contain multiple objects with each belonging to a distinct
category.

In this work, we address these two challenges in in-the-wild image retrieval by
proposing to \emph{use both a sketch and a text query as input}.
The (optional) input sketch is treated as a supplement to the text query to
provide more information that may be difficult to express with text (e.g.,
positions of multiple objects, object shapes). In particular, we do not require the input
sketch to be drawn well. As will be seen in our results, when
coupled with a text query, an extra input sketch (even a poorly drawn one) can
help considerably narrow down the set of candidate images, leading to an increase in
retrieval recall.
We show that both input modalities
can complement each other in a manner that cannot be easily achieved by either
one alone.  To illustrate this point concretely, two  example queries can be
found in \Cref{fig:teaser} where dropping one input modality would make it
difficult to retrieve the same set of images.
This idea directly addresses the first challenge of in-the-wild SBIR -- sketch ambiguity.

\begin{figure*}[t]
  \centering
  \includegraphics[width=4.5in]{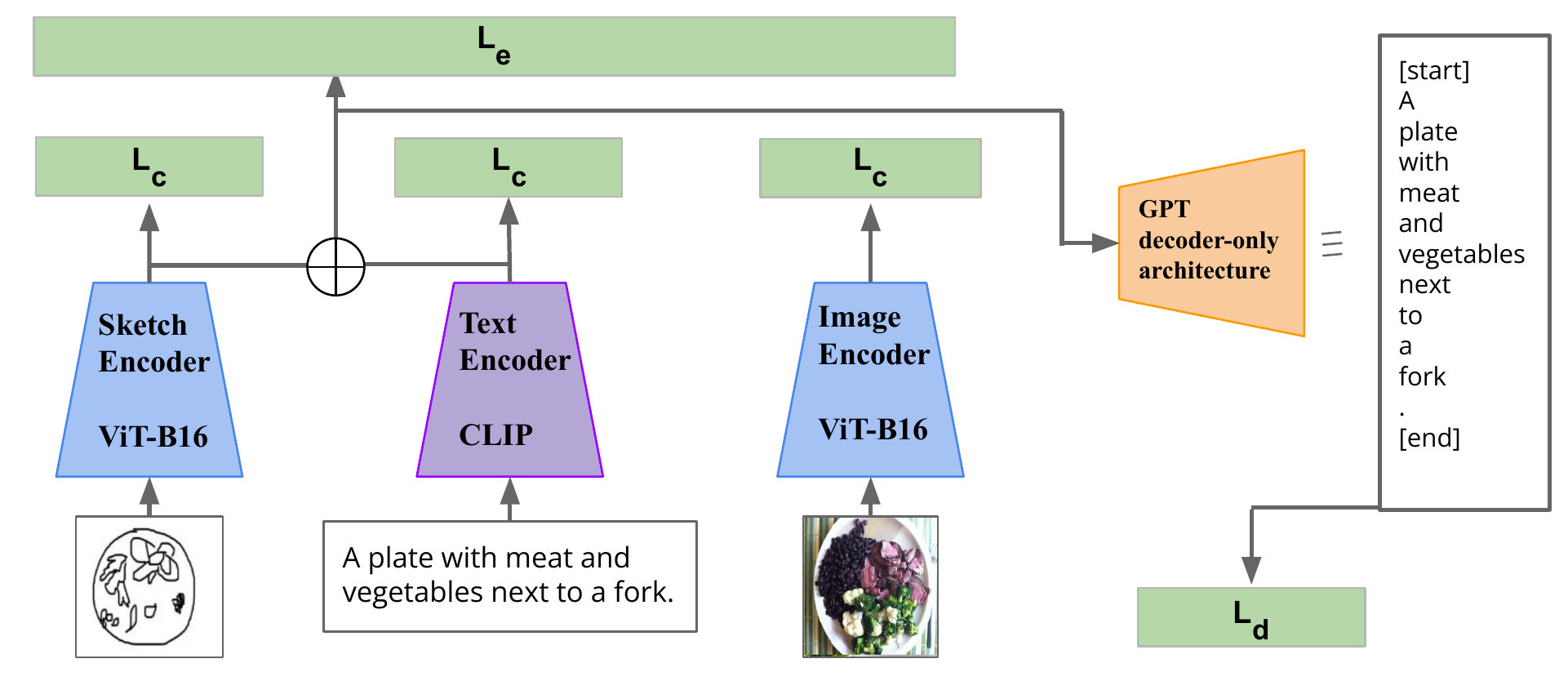}
 \caption{Overview of our model and the training pipeline.
 ViT is the vision transformer \cite{dosovitskiy2020vit}, and
 $\oplus$ represents an element-wise addition.
 We extend CLIP \cite{clip-radford21a} to incorporate additional sketch input from user. We add two auxiliary tasks: multi-label classification ($L_c$) and caption generation ($L_d$). Our objective losses are explained in \Cref{sec:obj}.
 We encourages the network to learn a discriminative representation to simultaneously distinguish at the instance level and at the category level, both of which are crucial for a successful retrieval.
 }
  \label{fig:diag}
\end{figure*}

Combining two input modalities is made possible with our proposed similarity
scoring model, \emph{\ourmethod{}}, which follows the late-fusion dual-encoder
approach, and allows efficient retrieval (see \Cref{fig:diag} for the model and
our training pipeline). Our proposed training objective comprises 1) an
embedding loss (to learn a shared embedding space for text, sketch, and image);
2) a multi-label classification loss
(to allow the model to recognize objects)
;
and 3) a caption generation loss (to encourage a strong correspondence between
the learned joint embedding and text description).
Crucially, training our model only requires synthetically generated sketches, not
human-drawn ones. These sketches are generated from the target images, and are
further transformed by appropriate augmentation procedures (e.g., random affine
transformation, dropout) to provide robustness to ambiguity in the input sketch.
The ability of our model to make use of synthetically generated sketches during
training addresses the second challenge of in-the-wild SBIR -- lack of training data.
We show that our model is robust to sketches with missing strokes
(\Cref{tab:sketch_complexity}), and is able to generalize and operate on
human-drawn input sketches (\Cref{fig:robustness}).

 Our contributions are as follows.
\begin{enumerate}
  \item We present \ourmethod{} (Text And SKetch transformer), a scalable, end-to-end trainable model
  for image retrieval using a text description and a sketch as input.
  %
  \item We collect 5000 hand-drawn sketches for images in the test set of COCO \cite{coco2018}, a commonly used dataset to benchmark image retrieval methods.
  The collected sketches are available at \projecturl{}.
  %
  \item We empirically demonstrate (in \Cref{sec:results}) that using an input sketch (even with a poorly drawn one) in addition to text
  helps increases retrieval recall, compared to the traditional TBIR where only a text description is used as input.
\end{enumerate}



\section{Related Work}
\label{sec:related_work}

\paragraph{Sketch Based Image retrieval (SBIR)}
There are several works that tackle SBIR
\cite{sketchy2016,pandey2020stacked,yu2016sketch,dutta2019semantically,Pang_2019_CVPR}.
The works of
\cite{pandey2020stacked,Dey_2019_CVPR,tursun2021efficient,liu2019semantic,dutta2019semantically} consider zero-shot SBIR where retrieving from unseen
categories is the focus. Zhang et al. \cite{zhang2016sketchnet} tackles SBIR
using only weakly labeled data.
Fine-grained SBIR of a restricted set of images (e.g., from a sigle category) is studied in
\cite{Pang_2019_CVPR,song2017deep,pang2017cross}.
An interactive variant of SBIR was considered in \cite{collomosse2019livesketch} where images are retrieved on-the-fly as an input sketch is being drawn. They
 propose a form of sketch completion system for SBIR based on user's choice of image cluster to refine the retrieval results.
Bhunia et al. \cite{bhunia2020sketch} present a reinforcement learning based pipeline for SBIR and allows a retrieval even before the sketch is completed.
Being able to retrieve relevant images even with incomplete sketch information is an important aspect of SBIR.
In this work, we achieved this by supplementing the (incomplete) sketch with an input text description.

\paragraph{Text-based image retrieval (TBIR)}
The problem of retrieving images based on a text description is closely related to the ability to learn a good representation between these two different domains.
Previous successes in TBIR relied on the cross attention mechanism (a specific form of early fusion) to  learn the similarity between text and images \cite{zhang2020context,lee2018stacked,li2020oscar}.
It is generally believed that, compared to a late fusion model, an early fusion model can offer more capacity for modeling a complex scoring function owing to its non-factorized form \cite{YanJinLinGuo2020,AlbLinColRei2019}.
Interestingly however the recently proposed CLIP \cite{clip-radford21a}, a late fusion model, is able to achieve comparable TBIR results on COCO to existing early fusion models.
This feat is impressive and is presumably due to the use of a large proprietary dataset of text-image pairs crawled from the Internet.
The representation learned by CLIP is incredibly rich, and correlates well with human perception on image-text similarity as observed in \cite{hessel2021clipscore}.

Similar to CLIP, ALIGN \cite{jia2021align} also proposes a similar contrastive training scheme on a large-scale data.
Fine-tuning from a pre-trained network in ALIGN is able to achieve state-of-the-art (SOTA) accuracy of 59.9\% in retrieving a single correct image based on text description in the COCO image retrieval benchmark (karphathy split).\footnote{More recently, Microsoft team also presented T-Bletchley \cite{tbletchley} with a similar pipeline. 
}
In our work, we opt to use the pre-trained CLIP  network to initialize our network for training.

\paragraph{Image retrieval with multi-modal queries} Using text and image as
queries for image retrieval has been explored extensively in the literature, for
example as a textual instruction for a desired modification of the query image
\cite{tautkute2019deepstyle,47874,han2017automatic,chen2020learning,dong2020using}.
 Jia et al \cite{jia2021align}  demonstrate that it is possible to directly combine representations of the two input modalities (text and image in this case) with a simple element-wise addition, for the purpose of image retrieval with maximum inner product search.
In our work, we take this insight and combine the representation of sketch and text inputs in the same way.
Using a sketch as input which is further supplemented by additional information (such as a classification label) has also been studied in Sketch2Tag \cite{wang2012sketch} and Sketch2Photo \cite{10.1145/1618452.1618470}.
Instead of sketch, \cite{changpinyo2021telling} explores using fine grain association between mouse trace and text description for image retrieval.
Surprisingly, combining sketch and full text description as queries for image retrieval has been relatively under explore despite its potential usefulness.
\cite{dey2018learning,han-schlangen-2017-draw,BMVC2017_45} explore the benefit of training with both sketch and text inputs for the purpose of improving image retrieval performance for each modality separately.
Only \cite{BMVC2017_45} demonstrate examples of retrieving with both sketch and text inputs by linearly combining their score during test time.
Note that unlike ours, the proposed loss function of \cite{BMVC2017_45} does not explicitly consider interactions between sketch and text inputs, since the goal is to simply improve retrieving with each modality separately.
Further to find the similarity between a sketch and a given image, the proposed method of \cite{BMVC2017_45} requires that the image be first converted to a sketch via edge detection. This conversion would incur a loss of information such as the color.
The proposed method also requires a conversion of image into sketch via edge detection, and would lose color information which is not ideal.
While \cite{BMVC2017_45} is one of the most related works to our work, we are unfortunately not able to obtain the data used, or the trained model for a quantitative comparison.


\section{Proposal: \ourmethod}
\label{sec:method}
\ourmethod{} can be considered an extension of the training pipeline proposed in CLIP \cite{clip-radford21a}.
We start with the same network architectures and contrastive learning as in CLIP, but modify them and add additional loss terms to allow the use of both a sketch and an input text query.
Specifically, we include two additional auxiliary tasks: multi-label classification and caption generation. Our motivation is to improve the learned embedding space so as to achieve the following goals.
\begin{enumerate}
    \item Discriminate between the positive and negative pairs;
    \item Distinguish objects of different categories; and
    \item Contain sufficient information to reconstruct the original text caption from embeddings of an image and its sketch.
\end{enumerate}
We achieve these goals via CLIP's symmetric cross entropy, multi-label classification objective, and caption generation objective, respectively.
While these three objectives appear to seek conflicting goals (e.g., the classification loss might encourage discarding class-invariant information, whereas the image captioning loss would suffer less if all describable details are kept), we observe that  combining them with appropriate weights can lead to gains in performance for image retrieval as shown in \Cref{tab:ablation}.

We start by describing our model and training pipeline in \Cref{sec:pipeline}, and  describe the three aforementioned loss terms in \Cref{sec:obj}.

\subsection{Model and Training Pipeline}
\label{sec:pipeline}
Our pipeline is summarized in  \Cref{fig:diag}.
Each input query consists of 1) a hand-drawn sketch, and 2) a text description of desired target images.
We use the same image and text encoder architecture as described in CLIP \cite{clip-radford21a} in order to leverage networks that have been the pre-trained with  large-scale training data.
This choice is in accord with the common practice of using, for instance, an ImageNet  pre-trained network for various vision tasks.
We use CLIP's publicly available pre-trained model ViT-B/16, in which the image encoder is based on the Vision Transformer \cite{dosovitskiy2020vit} which we found to be the best performer for our task.
For the details of the network architecture of each encoder that we use, please refer to CLIP \cite{clip-radford21a} and ViT \cite{dosovitskiy2020vit}.

There are three encoders in our pipeline for: 1) input sketch, 2) input text description, and 3) a candidate retrieved image.
The output embeddings from the sketch and the text encoders are  combined together
and used for contrastive learning with the image embedding as the target. We explore several options for combining features in \Cref{sec:ablation}.
Since images and sketches are in the same domain (visual), we use the same architecture for them (ViT-B/16 pre-trained on CLIP).
We also found that the performance is much better when sharing weight parameters across the sketch and the image encoders.
Yu et al.  \cite{yu2016sketch} also observed similar results where the Siamese network performs significantly better than heterogeneous networks for the SBIR task.
It was speculated that weight sharing is advantageous because of relatively small training datasets. This explanation may also hold in our case as well given the complexity of our task.
For classification, we feed each embedding to two additional fully connected layers with ReLU activation.

Additionally, we also train a captioning generator from the embeddings of sketches and images using a transformer based text decoder.
This decoder is an autoregressive language model similar to GPT decoder-only architecture \cite{radford2019language}. We use absolute positional embedding, with six stacks of decoder blocks, each with eight attention heads. More details can be found in the supplementary materials.

\subsection{Objective Function}
\label{sec:obj}
Our objective function consists of three main components: symmetric cross entropy, asymmetric loss for multi-label classification, and  auxiliary caption generation.

\titledpar{Embedding Loss ($L_e$)}
To learn a shared embedding space for text, image and sketch, we follow the contrastive learning objective from CLIP \cite{clip-radford21a}, which use a form of \emph{InfoNCE Loss} as originally proposed in \cite{oord2018representation} to learn to match the right image-text pairs as observed in the batch.
This proxy task is accomplished via a symmetric cross entropy loss over all possible pairs in each batch, effectively maximizing cosine similarity of each matching pair and minimizing it for  non-matching ones.
We add the sketch as an additional query, and replace the text embedding in CLIP with our combined embedding constructed by summing the text and sketch embeddings.

\titledpar{Classification Loss ($L_c$)}
For classification, we consider this as a multi-label classification problem as each image can belong to multiple categories.
We follow a common practice in multi-label classification, which frames the problem as a series of many binary classification problems.
We use object annotation available in the datasets as ground truth. Specifically, we use Asymmetric Loss For Multi-Label Classification (ASL Loss), proposed in \cite{benbaruch2020asymmetric}. The loss is designed to help alleviate the effect of the imbalanced label distribution in multi-label classification.

\titledpar{Auxiliary caption generation (Decoder Loss, $L_d$)}
For caption generation, the decoder attempts to predict the most likely token given the accumulated embedding and the previous tokens.
The decoded output tokens are then compared with the ground truth sentences via the cross entropy loss
$\sum_t^T{\log(p_{t}|p_1, \ldots, p_{t-1})},$
where $T$ is the maximum sequence length.

Our final objective is given by a weighted combination of all the loss terms. We refer to supplementary materials for our hyperparameter choices.

\subsection{Sketch Generation and Data Augmentation}
\label{sec:augmentation}
During training, we synthetically generate sketches using the method proposed in \cite{LIPS2019}.
In general the method produces drawings similar to human sketches.
The synthesized sketches are however in exact alignment with their source images, which would not be the case in sketches drawn by humans.
To achieve invariance to small misalignment, we further apply a random affine transformation on the synthetic sketch as an augmentation.
We also apply a similar transformation to the image (with different random seeds).
Introducing this misalignment is crucial for the network to generalize to hand-drawn input sketches at test time.

To help deal with partial sketches at test time, we also randomly occlude parts of each sketch. We randomly replace black strokes with white pixel. The completion level of each synthesized sketch in the training set is between 60\% to 100\%

\paragraph{Evaluation with sketch-text-image tuples}
Since our approach retrieves images with sketches and text, evaluating our approach naturally requires an annotated dataset where each record contains a hand-drawn sketch, a human-annotated text description, and a source image.
SketchyCOCO \cite{Gao_2020_CVPR} fits this description and is a candidate dataset.
However, the dataset was constructed by having the sketches drawn first based on categories.
The sketch for each category was then pasted into their supposed areas based on incomplete annotation (not all objects were annotated).
As a result, the sketch in each record may poorly represent the image because 1) each category contains the same sketch, 2) the choice of which object to draw is based on categories, rather than human judgement of what is salient in each particular image.
More related is the dataset mentioned in \cite{BMVC2017_45} which provides 1,112 matching sketch/image/text of shoes.
However, the dataset is not yet available at the time of writing.
Also worth mentioning is \cite{PontTuset_eccv2020}, which also proposes a dataset containing synchronized annotation between text description and the associated location (in a form of mouse trace) in the image. However, we argue that a line drawing sketch represents more than just the location of the objects; even a badly drawn sketch can provide information such as shape, details, or even relative scale.
Owing to lack of an appropriate evaluation dataset, we construct a new benchmark by collecting hand-drawn sketches for images in the COCO image retrieval benchmark.
These images are in the COCO 5k split from \cite{karpathy2015deep}, which is widely used to evaluate text based image retrieval methods.
As part of our contributions, the collected data will be made publicly available.
We describe the sketch collection process in the next section.

\subsection{Data Collection: Sketching from Memory}
\label{sec:data_collect}

We collect sketches for COCO 5k via Amazon Mechanical Turk crow sourcing platform (AMT).
We follow the split from \cite{karpathy2015deep}, which has been widely used as benchmark for text based image retrieval. The test set contains 5000 images which are disjoint from the training split.

To solicit a sketch, we first show the participant (Turker) the target image for 15 seconds, before replacing it with a noise image.
This is to mimic the typical scenario at deployment time where we there is no concrete reference image, but instead only a mental image of a retrieval target.
The participant is then asked to draw a sketch from memory.
In the instructions, we ask the participants to draw as if they are explaining the image to a friend but using a sketch.
We do not put any restrictions on how the sketch has to be drawn, other than no shading.
The participants are free to draw any parts of the image they think are important and distinctive enough to be included in the sketch.
The sketches are collected in SVG (a vector format), and contain information of all individual strokes.
In experiments, we use this information to randomly drop individual strokes to test the robustness of our approach to incomplete input sketches (see \Cref{tab:sketch_complexity}).
As a sanity check, we also ask the participants to put one or two words describing the image in an open-ended fashion.
%

From the results, we manually filter out sketches that are not at all representative of the target image (e.g., empty sketch, random lines, wrong object).
Our only criteria is that each sketch has to be recognizable as describing the target image (even if only remotely recognizable).
Our goal is not to collect complete or perfect sketches, but to collect in-the-wild sketches, which may be poorly drawn, that can be used along side the text description to explain image.


\section{Results and Discussion}
\label{sec:results}

For quantitative evaluation, we calculate Recall@K, which is the fraction of times that the target image is included in the top-K retrieved images.
We evaluate on COCO \cite{coco2018}; a commonly used datasets for language based image retrieval.
We use the same data split as proposed in \cite{karpathy2015deep}.
Specifically, COCO's evaluation set contains 5,000 images.
Each image is annotated with multiple captions, and we additionally add a sketch to each image.
For COCO, we collect hand drawn sketch as described in \ref{sec:data_collect}.

\begin{figure*}[t]
  \centering
  \includegraphics[width=\linewidth]{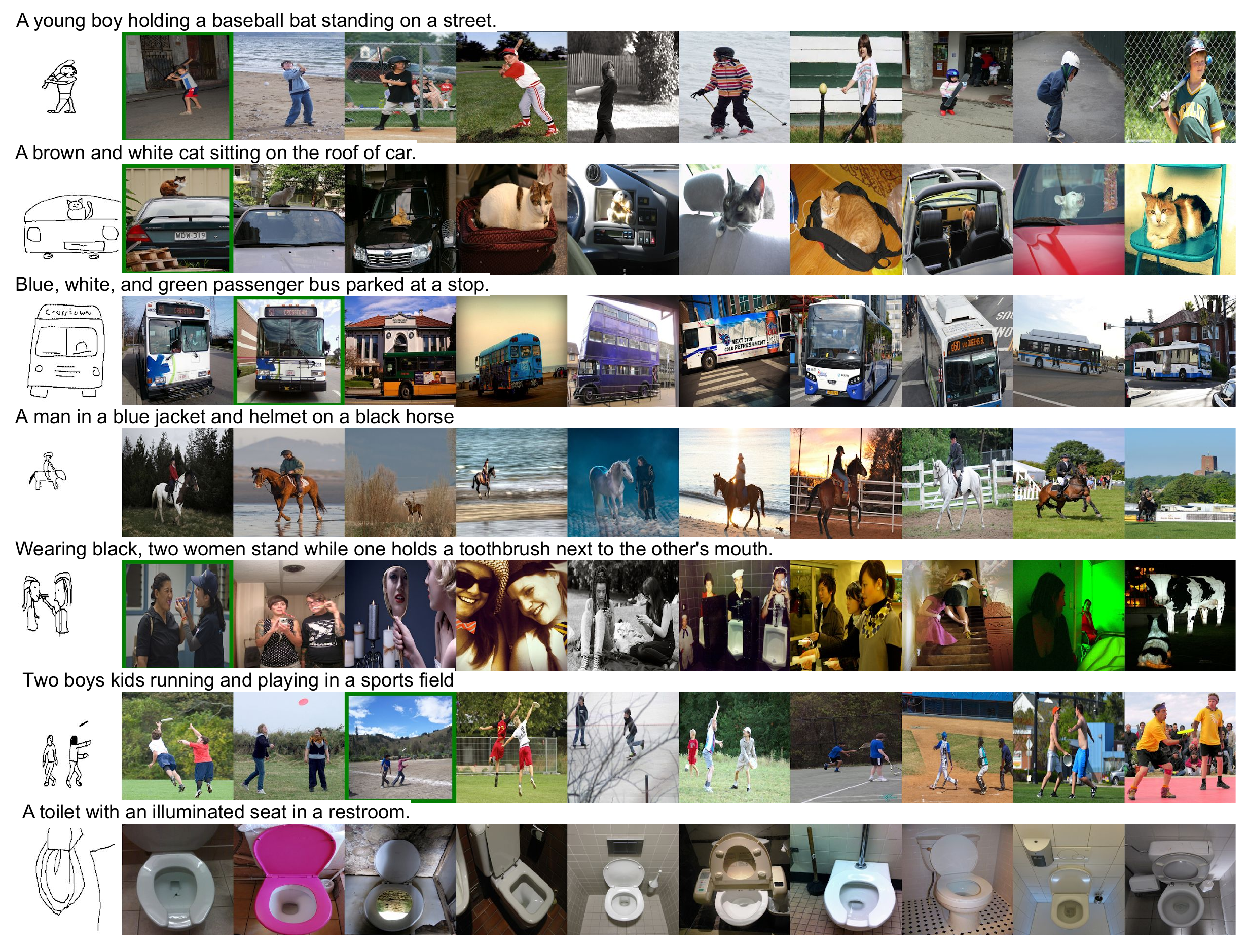}
 \caption{
 Example retrieved images from \ourmethod{}, randomly selected from our benchmark.
Each query consists of a text description (shown at the top of each block, and an input sketch (at the top left of each block).
 In each case, the image that forms a matching pair to the sketch is highlighted with a green border.
 See \Cref{sec:data_collect} for details on how we  collect human drawn sketches for images in the evaluation set.
  }
  \label{fig:qual_res}
\end{figure*}

\paragraph{Implementation details}

We use Adam \cite{KinBa2014} as the optimization method to train the model.
Training hyperparameters are set in accordance with Open Clip \cite{ilharco_gabriel_2021_5143773} with the initial learning rate set to $10^{-5}$.
To provide robustness to incomplete input, for each (text query, sketch, image) training tuple, we set a 20\% probability to drop either the sketch (replaced with a white image) or the text query (replaced with an empty text).
We demonstrate the robustness of our model
to incomplete input in \Cref{sec:robustness}.
Code to reproduce our results will be made publicly available.



\begin{table*}[t]
\caption{ Recall@\{1, 5, 10\} of state-of-the-art methods on COCO dataset.
We use our collected hand-drawn sketches for evaluation.
Among ALIGN, and CLIP, only CLIP has released pre-trained model.
Our \ourmethod{} is initialized for training with this CLIP model.
}
\centering
\begin{tabular}{p{2.5cm}p{4.1cm}p{2.2cm}lll}
\cmidrule{1-6}
Type & Method &   Input &       R@1 &        R@5 &       R@10 \\
\cmidrule{1-6}
Early fusion & Uniter \cite{chen2020uniter}  &  Text &      0.529 &      0.799 &      0.880 \\
Early fusion & OSCAR \cite{li2020oscar}&   Text &     0.575 &      0.828 &      0.898 \\
Dual encoders & ALIGN \cite{jia2021align} (fine-tuned)  &  Text &     0.599 &      0.833 &      0.898 \\
Dual encoders & CLIP \cite{clip-radford21a} (fine-tuned) &  Text &     0.518 &      0.811 &      0.891 \\
\cmidrule{1-6}
Dual encoders & \textbf{\ourmethod{}} (ours) &  (Text, Sketch) &    \textbf{0.609} &      \textbf{0.847} &      \textbf{0.917} \\
\cmidrule{1-6}
\end{tabular}
\label{tab:quan_res}
\end{table*}

For evaluation, our model \ourmethod{} is initialized with the publicly released CLIP model  (ViT-16) \cite{clip-radford21a}, a text-based image retrieval model.
\Cref{tab:quan_res} compares Recall@\{1, 5, 10\} of our method and that of current state-of-the-art approaches on text based image retrieval: Uniter, OSCAR, and ALIGN as reported in \cite{chen2020uniter,jia2021align,li2020oscar}, respectively.
To provide a fairer comparison, we further finetune the publicly released CLIP model (ViT-16) on COCO.\footnote{Note that this baseline makes use of our
best-effort implementation and training. We do not have access to the official
training code and there is no reported result on using a fine-tuned CLIP on an
image retrieval dataset.}
We observe that our method is able to achieve a considerably higher recall than other
methods, and CLIP in particular, owing in part to the use of a sketch as a supplemental input to text.
Our method can retrieve the correct image
with recall@1 of 60.9\%, compared to using text alone, which achieves 51.8\% on the same
CLIP architecture (ViT-16).

At the time of writing, models for ALIGN and T-Bletchley have not been released. Both methods have been shown to perform better than CLIP on text based image retrieval, and without any early fusion of text and image. We note that both ALIGN and T-Bletchley can be used as part of our framework by simply replacing the pre-trained encoders for text and image.




\begin{table}[t]
\caption{Results from our abalation study as described in \Cref{sec:ablation}. 
We construct variants of our proposed method by selectively including only some training losses.
These are denoted by $L_e$, $L_e + L_c$, and $L_e + L_c + L_d$ (see \Cref{sec:obj}).
Sketch and text inputs are combined by adding their embeddings. 
For ``Feature max'' and ``Feature concat'', the embeddings are combined by coordinate-wise maximum, and concatenation, respectively, and the full objective is used.
}
\centering
\begin{tabular}{p{3.8cm}lll}
\midrule
                                               Method &          R@1 &        R@5 &       R@10 \\
\midrule
 CLIP (zero shot) &       0.378 &      0.624 &      0.722 \\
\hline
 Ours: $L_e$ &     0.493 &      0.748 &      0.836 \\
 Ours: $L_e + L_c$  &      0.509 &      0.764 &      0.850 \\
Ours: $L_e + L_c + L_d$ &       0.527 &      0.778 &      0.862 \\

\hline
 Ours: Feature max&      0.443 &      0.704 &      0.804 \\
 Ours: Feature concat&      0.357 &      0.650 &      0.768 \\
\hline
 Ours (final) &      \textbf{0.609} &      \textbf{0.847} &      \textbf{0.917} \\
\bottomrule
\end{tabular}

\label{tab:ablation}
\end{table}

\subsection{Ablation Study}
\label{sec:ablation}

In this section, we seek to understand the effect of each of our proposed loss functions ($L_e, L_c, L_d$), as described in \Cref{sec:method}.
Baselines used as part of this ablation study are:
\begin{itemize}
    \item
\textbf{CLIP zero shot}. Recall@K that has been reported in \cite{clip-radford21a} for zero-shot image retrieval.
We note that their best model for the reported results (ViT-L/14@336px) is not available publicly.
\end{itemize}
%
\begin{itemize}
    \item
\textbf{Ours: $L_e$} (ablated \ourmethod{}). We start adding a sketch as an additional query, as shown in \Cref{fig:diag}. The only objective  in this baseline is to correctly classify the correct matching pair between the query (sketch+text) and the image via \emph{symmetric cross entropy loss}.
    \item
\textbf{Ours: $L_e + L_c$} (ablated \ourmethod{}). In this baseline, we add the multi-label classification loss term in the objective.
    \item
\textbf{Ours: $L_e + L_c + L_d$} (ablated \ourmethod{}). In this baseline, we add both classification loss and decoder loss
\end{itemize}
For the above three baselines, text and sketch embeddings are combined by adding them, as described in \Cref{fig:diag}.
We further compare two additional ways to combine sketch and text embeddings:  coordinate-wise maximum, and concatenation:
\begin{itemize}
    \item
\textbf{Ours: Feature max} (ablated \ourmethod{}). Embeddings from sketch and text are combined using element-wise max.
\item
\textbf{Ours: Feature concatenate} (ablated \ourmethod{}). Embeddings from sketch and text are concatenated, and projected into the same dimension as embedding from image.
\end{itemize}
We use  the full objective  (i.e., $L_e + L_c + L_d$) for the above two variants.

\begin{itemize}
    \item
\textbf{Ours (final)} This is our complete model trained with the full objective ($L_e + L_c + L_d$). We augment training sketches and images with random affine transformation, randomly remove parts of each sketch (as describe in \cref{sec:method}) and train for 50 epochs.
\end{itemize}
Except for the final model, we train each baseline for 10 epoches. For ablation study, we simplify the training by only performing simple augmentation (random cropping and flipping).

\Cref{tab:ablation} reports recalls of each baselines. Both $L_c$ and $L_d$ further improve the retrieval performance compare to our baseline with only embedding loss ($L_e$). Surprisingly, our experiment also show that a simple element wise addition leads to the best performance compare to element wise max, and concatenation. We hypothesize that direct combination of the sketch and text embedding implicitly helps with feature alignment, compare to concatenate them.


\begin{figure*}[t]
  \centering
  \includegraphics[width=\linewidth]{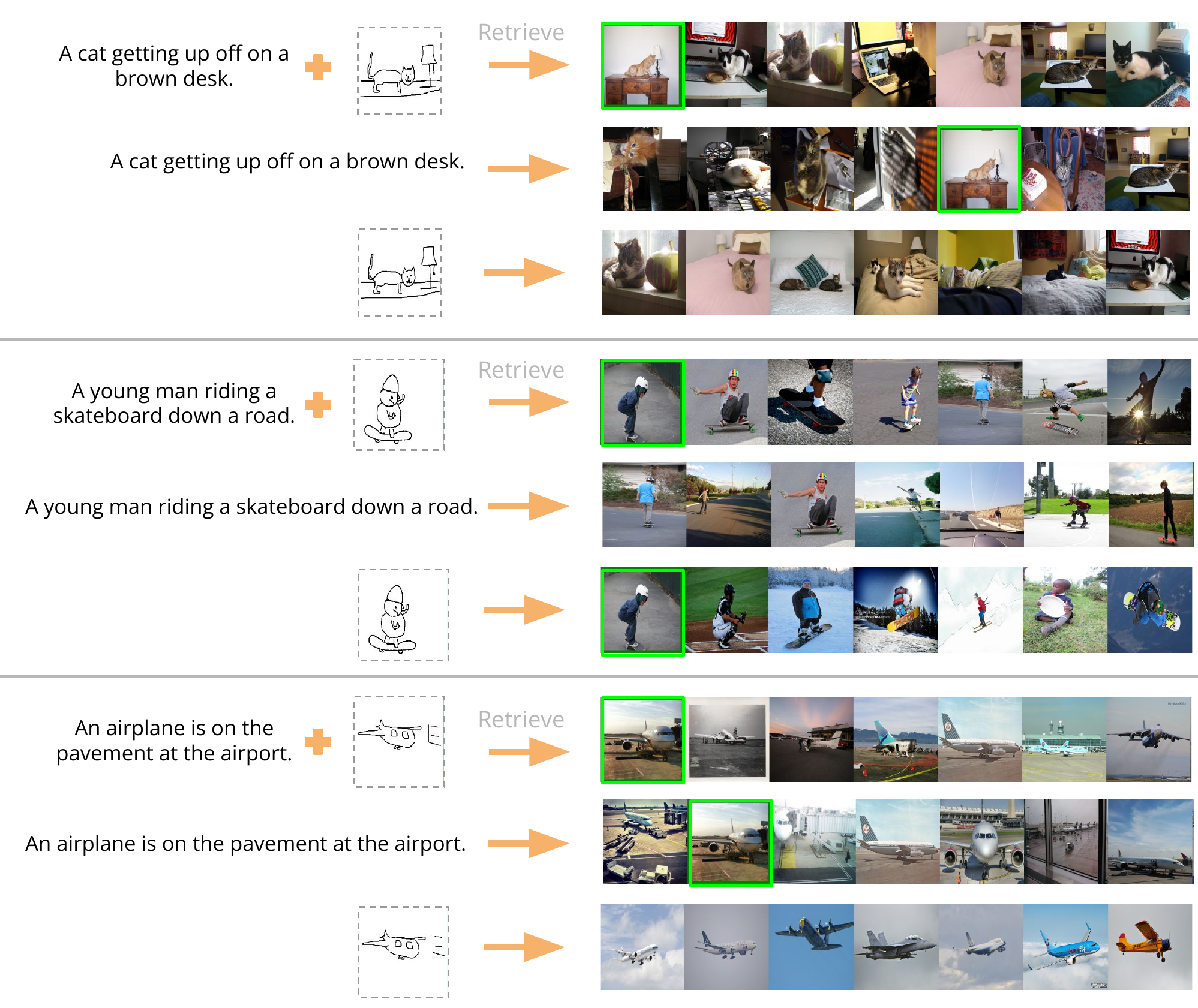}
 \caption{Some retrieved images by our model when using only a sketch query, only a text query, and with both queries at the same time.
 Our model is robust to missing input: when only one input modality is present, it can still retrieve relevant image, albeit suboptimally. We observe that the presence of an input sketch clearly helps rank the most relevant image higher.}
  \label{fig:robustness}
\end{figure*}


\subsection{Robustness to Missing Input}
\label{sec:robustness}
One key benefit of our pipeline is the ability to retrieve image even when missing one of the input modalities (sketch or text).
We accomplish this by adding a query drop-out augmentation which replaces either the sketch or text with an empty sketch or an empty string.
This ensures the network can operate even when no sketch or text description is given.
\Cref{fig:robustness} shows the results breaking down into querying with sketch only, querying with text only, and querying with both modalities. We emphasize that sketch based image retrieval is particularly challenging for unconstrained, in-the-wild images with multiple objects and no fixed categories. In spite of that, our network can retrieve reasonably relevant results for sketch query alone, all without requiring any hand drawn sketch during training. 

Nevertheless, we find that using the sketch or text as a sole input is often
sub-optimal for image retrieval in this setup.
A combination of sketch and text can provide a more complete picture of the target
image, leading to an improve performance. For instance, in the first example of
\Cref{fig:robustness}, it can be difficult to draw a \emph{brown} desk with
sketch alone, but this can be included easily with text description.
\subsection{Sketch Complexity and Retrieval Performance}
\label{sec:sketch_complexity}


\begin{figure*}[t]
  \centering
  \includegraphics[width=0.9\linewidth,trim={0 0 0 94mm},clip]{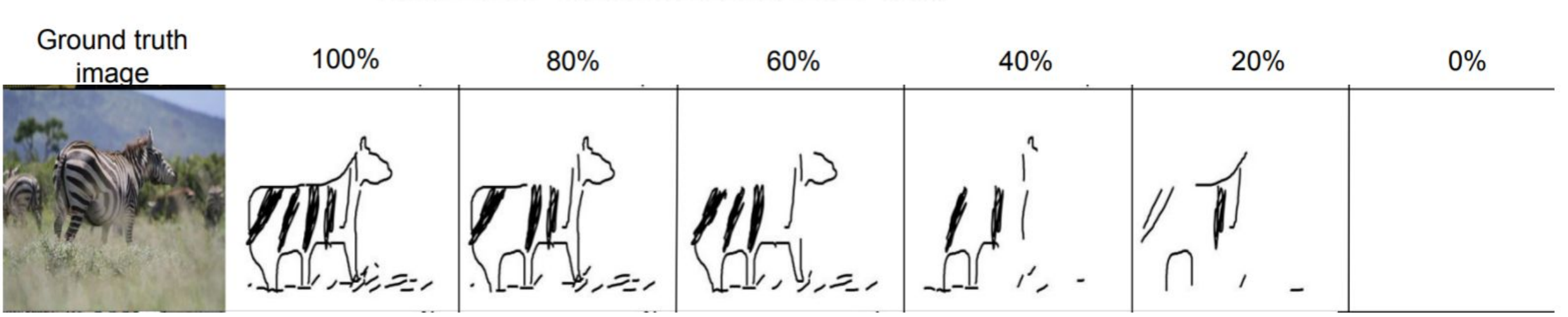}

%
\begin{tabular}{rcccccc}
  \cmidrule{1-7}
  Sketch completeness & 100\% \, & 80\% \, & 60\% \, & 40\% \, & 20\% \, & 0\% \\
  \cmidrule{1-7}
  R@1 & .615 & .609 & .602 & .582 & .515 & .458 \\
  R@5 & .883 & .876 & .870 & .843 & .833 & .776 \\
  R@10 & .947 & .926 & .940 & .923 & .916 & .870 \\
  \cmidrule{1-7}
\end{tabular}
\caption{
Recall@\{1, 5, 10\} of \ourmethod, averaged over 300 randomly sampled (text query, sketch, image) tuples in the COCO dataset.
For each test instance, we vary the completeness of the sketch, where completeness level $x\%$ is defined as the original sketch with only $x\%$ of all strokes randomly chosen.
Candidate retrievable images are the full COCO test set of 5k.
The completeness 0\% corresponds to using only text as input.
We observe that even with a small fraction of strokes retained (e.g., 20\%), there is gain in recall compared to using only text as input (i.e., 0\%  sketch completeness)
}
\label{tab:sketch_complexity}
\end{figure*}

In this section, we investigate the effect of varying the level of sketch details on the retrieval performance.
We  sample 300 sketches from our collected COCO hand-drawn sketches, and randomly subsample strokes to keep only 20\% to 100\%. Table \ref{tab:sketch_complexity} shows the recall performance on each level of sketch complexity.
We observe  a large gain near the lower end of the completeness level, with a
diminishing gain as the completeness level increases.
In particular, this suggests that even a poorly drawn sketch helps in retrieving more relevant images.
We observe that reducing the sketch completeness level from 100\% to 60\% only slightly decreases recall.


\subsection{On the Effect of Text Completeness}
In \Cref{sec:sketch_complexity}, we investigate the effect of an incomplete input sketch on the retrieval recall.
The goal of this section is to provide an analogous analysis on the effect of an incomplete text input.
We sample 300 records of from our collected COCO dataset.
Each record consists of three objects:  a sketch, a text description, and an image.
For each text description, we vary its degree of completeness by randomly subsampling $x\%$ (for a number of values of $x$) of the tokens (words), while keeping its corresponding sketch intact.
\Cref{tab:text_complexity} shows the resulted retrieval recall for each degree of text completeness.
We observe that compared to when no text query is given as an input (0\%), recall@1 increases from 0.099 to 0.316 when only 20\% of tokens are included.
The gain in recall diminishes as more tokens are added.

\begin{table}[t]
\caption{Retrieval performance of our model as measured by recall@\{1, 5, 10\} when both a complete sketches, and text queries are inputted to the model.
Words in each text query are sub-sampled to investigate the effect of incomplete input text on retrieval.
}
    \centering
  \begin{tabular}{rrrr}
  \toprule
  Text completeness &  ~  R@1 & ~  R@5 & ~ R@10 \\
  \midrule
          0\% & 0.099&0.195 &0.257 \\
          20\% & 0.316&0.545 &0.640 \\
          40\% & 0.357& 0.613&0.708 \\
          60\% & 0.446 & 0.710 & 0.813 \\
          80\% & 0.530 & 0.807 & 0.884 \\
          100\% & 0.607 & 0.866 & 0.930 \\
  \bottomrule
  \end{tabular}
    \label{tab:text_complexity}
\end{table}

\section{Limitations and Future Work}
\Cref{fig:qual_res} shows the potentials of our retrieval model. However,
it also reveals some examples of when our method fails to retrieve the target image. For example, in the forth row, the network is unable to retrieve the correct image (with blue jacket and black horse). This is likely due to the confusion with the color (both black and blue color are in the text description). While the network does place images with both black and blue color closer to the query embedding, it is unable to find the target image with black horse, and blue jacket within the top 10 results. This shows the current limitation of the model in understanding complicated text description.

Beyond these examples, we generally observe that the network suffers when the sketch is not representative of the target image.
For example, the difference in scale and location can contribute to incorrect retrieval results (sixth row of \Cref{fig:qual_res}).
Designing a model that is more tolerant to scale mismatch will be an interesting topic for future research.
On the text encoder, automatically augmenting the text query to be more concrete by leveraging text augmentation techniques also deserves further attention.


\clearpage

{\small
\bibliographystyle{splncs04}
\bibliography{egbib}

\begin{thebibliography}{10}
\providecommand{\url}[1]{\texttt{#1}}
\providecommand{\urlprefix}{URL }
\providecommand{\doi}[1]{https://doi.org/#1}

\bibitem{AlbLinColRei2019}
Alberti, C., Ling, J., Collins, M., Reitter, D.: Fusion of detected objects in
  text for visual question answering. arXiv preprint arXiv:1908.05054  (2019)

\bibitem{benbaruch2020asymmetric}
Ben-Baruch, E., Ridnik, T., Zamir, N., Noy, A., Friedman, I., Protter, M.,
  Zelnik-Manor, L.: Asymmetric loss for multi-label classification (2020)

\bibitem{bhunia2020sketch}
Bhunia, A.K., Yang, Y., Hospedales, T.M., Xiang, T., Song, Y.Z.: Sketch less
  for more: On-the-fly fine-grained sketch-based image retrieval. In:
  Proceedings of the IEEE/CVF Conference on Computer Vision and Pattern
  Recognition. pp. 9779--9788 (2020)

\bibitem{changpinyo2021telling}
Changpinyo, S., Pont-Tuset, J., Ferrari, V., Soricut, R.: Telling the what
  while pointing to the where: Multimodal queries for image retrieval. In:
  Proceedings of the IEEE/CVF International Conference on Computer Vision. pp.
  12136--12146 (2021)

\bibitem{10.1145/1618452.1618470}
Chen, T., Cheng, M.M., Tan, P., Shamir, A., Hu, S.M.: Sketch2photo: Internet
  image montage. ACM Trans. Graph.  \textbf{28}(5),  1--10 (dec 2009).
  \doi{10.1145/1618452.1618470}

\bibitem{chen2020learning}
Chen, Y., Bazzani, L.: Learning joint visual semantic matching embeddings for
  language-guided retrieval. In: Computer Vision--ECCV 2020: 16th European
  Conference, Glasgow, UK, August 23--28, 2020, Proceedings, Part XXII 16. pp.
  136--152. Springer (2020)

\bibitem{chen2020uniter}
Chen, Y.C., Li, L., Yu, L., Kholy, A.E., Ahmed, F., Gan, Z., Cheng, Y., Liu,
  J.: Uniter: Universal image-text representation learning. In: ECCV (2020)

\bibitem{collomosse2019livesketch}
Collomosse, J., Bui, T., Jin, H.: {Livesketch}: Query perturbations for guided
  sketch-based visual search. In: Proceedings of the IEEE/CVF Conference on
  Computer Vision and Pattern Recognition. pp. 2879--2887 (2019)

\bibitem{dey2018learning}
Dey, S., Dutta, A., Ghosh, S.K., Valveny, E., Llad{\'o}s, J., Pal, U.: Learning
  cross-modal deep embeddings for multi-object image retrieval using text and
  sketch. In: 2018 24th international conference on pattern recognition (ICPR).
  pp. 916--921. IEEE (2018)

\bibitem{Dey_2019_CVPR}
Dey, S., Riba, P., Dutta, A., Llados, J., Song, Y.Z.: Doodle to search:
  Practical zero-shot sketch-based image retrieval. In: The IEEE Conference on
  Computer Vision and Pattern Recognition (CVPR) (June 2019)

\bibitem{dong2020using}
Dong, H., Wang, Z., Qiu, Q., Sapiro, G.: Using text to teach image retrieval
  (2020)

\bibitem{dosovitskiy2020vit}
Dosovitskiy, A., Beyer, L., Kolesnikov, A., Weissenborn, D., Zhai, X.,
  Unterthiner, T., Dehghani, M., Minderer, M., Heigold, G., Gelly, S.,
  Uszkoreit, J., Houlsby, N.: An image is worth 16x16 words: Transformers for
  image recognition at scale. ICLR  (2021)

\bibitem{dutta2019semantically}
Dutta, A., Akata, Z.: Semantically tied paired cycle consistency for zero-shot
  sketch-based image retrieval. In: Proceedings of the IEEE/CVF Conference on
  Computer Vision and Pattern Recognition. pp. 5089--5098 (2019)

\bibitem{Gao_2020_CVPR}
Gao, C., Liu, Q., Xu, Q., Wang, L., Liu, J., Zou, C.: Sketchycoco: Image
  generation from freehand scene sketches. In: Proceedings of the IEEE/CVF
  Conference on Computer Vision and Pattern Recognition (CVPR) (June 2020)

\bibitem{han-schlangen-2017-draw}
Han, T., Schlangen, D.: Draw and tell: Multimodal descriptions outperform
  verbal- or sketch-only descriptions in an image retrieval task. In:
  Proceedings of the Eighth International Joint Conference on Natural Language
  Processing (Volume 2: Short Papers). pp. 361--365. Asian Federation of
  Natural Language Processing, Taipei, Taiwan (Nov 2017),
  \url{https://aclanthology.org/I17-2061}

\bibitem{han2017automatic}
Han, X., Wu, Z., Huang, P.X., Zhang, X., Zhu, M., Li, Y., Zhao, Y., Davis,
  L.S.: Automatic spatially-aware fashion concept discovery. In: ICCV (2017)

\bibitem{hessel2021clipscore}
Hessel, J., Holtzman, A., Forbes, M., Bras, R.L., Choi, Y.: Clipscore: A
  reference-free evaluation metric for image captioning (2021)

\bibitem{flickr15k}
Hu, R., Collomosse, J.: A performance evaluation of gradient field hog
  descriptor for sketch based image retrieval. Comput. Vis. Image Underst.
  \textbf{117}(7),  790--806 (jul 2013). \doi{10.1016/j.cviu.2013.02.005},
  \url{https://doi.org/10.1016/j.cviu.2013.02.005}

\bibitem{ilharco_gabriel_2021_5143773}
Ilharco, G., Wortsman, M., Carlini, N., Taori, R., Dave, A., Shankar, V.,
  Namkoong, H., Miller, J., Hajishirzi, H., Farhadi, A., Schmidt, L.: Openclip
  (Jul 2021). \doi{10.5281/zenodo.5143773},
  \url{https://doi.org/10.5281/zenodo.5143773}

\bibitem{jia2021align}
Jia, C., Yang, Y., Xia, Y., Chen, Y.T., Parekh, Z., Pham, H., Le, Q.V., Sung,
  Y., Li, Z., Duerig, T.: Scaling up visual and vision-language representation
  learning with noisy text supervision. arXiv preprint arXiv:2102.05918  (2021)

\bibitem{BMVC2017_45}
Jifei~Song, Yi-zhe~Song, T.X., Hospedales, T.: Fine-grained image retrieval:
  the text/sketch input dilemma. In: Tae-Kyun~Kim, Stefanos~Zafeiriou, G.B.,
  Mikolajczyk, K. (eds.) Proceedings of the British Machine Vision Conference
  (BMVC). pp. 45.1--45.12. BMVA Press (September 2017). \doi{10.5244/C.31.45}

\bibitem{karpathy2015deep}
Karpathy, A., Fei-Fei, L.: Deep visual-semantic alignments for generating image
  descriptions (2015)

\bibitem{KinBa2014}
Kingma, D.P., Ba, J.: Adam: A method for stochastic optimization. arXiv
  preprint arXiv:1412.6980  (2014)

\bibitem{lee2018stacked}
Lee, K.H., Chen, X., Hua, G., Hu, H., He, X.: Stacked cross attention for
  image-text matching. arXiv preprint arXiv:1803.08024  (2018)

\bibitem{LIPS2019}
Li, M., Lin, Z., M\v~ech, R., , Yumer, E., Ramanan, D.: Photo-sketching:
  Inferring contour drawings from images. WACV  (2019)

\bibitem{li2020oscar}
Li, X., Yin, X., Li, C., Hu, X., Zhang, P., Zhang, L., Wang, L., Hu, H., Dong,
  L., Wei, F., Choi, Y., Gao, J.: Oscar: Object-semantics aligned pre-training
  for vision-language tasks. ECCV 2020  (2020)

\bibitem{coco2018}
Lin, T., Maire, M., Belongie, S.J., Bourdev, L.D., Girshick, R.B., Hays, J.,
  Perona, P., Ramanan, D., Doll{\'{a}}r, P., Zitnick, C.L.: Microsoft {COCO:}
  common objects in context. CoRR  \textbf{abs/1405.0312} (2014),
  \url{http://arxiv.org/abs/1405.0312}

\bibitem{liu2019semantic}
Liu, Q., Xie, L., Wang, H., Yuille, A.L.: Semantic-aware knowledge preservation
  for zero-shot sketch-based image retrieval. In: Proceedings of the IEEE/CVF
  International Conference on Computer Vision. pp. 3662--3671 (2019)

\bibitem{nilsback2008automated}
Nilsback, M.E., Zisserman, A.: Automated flower classification over a large
  number of classes. In: 2008 Sixth Indian Conference on Computer Vision,
  Graphics \& Image Processing. pp. 722--729. IEEE (2008)

\bibitem{oord2018representation}
Oord, A.v.d., Li, Y., Vinyals, O.: Representation learning with contrastive
  predictive coding. arXiv preprint arXiv:1807.03748  (2018)

\bibitem{pandey2020stacked}
Pandey, A., Mishra, A., Verma, V.K., Mittal, A., Murthy, H.: Stacked
  adversarial network for zero-shot sketch based image retrieval. In:
  Proceedings of the IEEE/CVF Winter Conference on Applications of Computer
  Vision. pp. 2540--2549 (2020)

\bibitem{Pang_2019_CVPR}
Pang, K., Li, K., Yang, Y., Zhang, H., Hospedales, T.M., Xiang, T., Song, Y.Z.:
  Generalising fine-grained sketch-based image retrieval. In: Proceedings of
  the IEEE/CVF Conference on Computer Vision and Pattern Recognition (CVPR)
  (June 2019)

\bibitem{pang2017cross}
Pang, K., Song, Y.Z., Xiang, T., Hospedales, T.M.: Cross-domain generative
  learning for fine-grained sketch-based image retrieval. In: BMVC. pp. 1--12
  (2017)

\bibitem{PontTuset_eccv2020}
Pont-Tuset, J., Uijlings, J., Changpinyo, S., Soricut, R., Ferrari, V.:
  Connecting vision and language with localized narratives. In: ECCV (2020)

\bibitem{clip-radford21a}
Radford, A., Kim, J.W., Hallacy, C., Ramesh, A., Goh, G., Agarwal, S., Sastry,
  G., Askell, A., Mishkin, P., Clark, J., Krueger, G., Sutskever, I.: Learning
  transferable visual models from natural language supervision. In: Meila, M.,
  Zhang, T. (eds.) Proceedings of the 38th International Conference on Machine
  Learning. Proceedings of Machine Learning Research, vol.~139, pp. 8748--8763.
  PMLR (18--24 Jul 2021),
  \url{https://proceedings.mlr.press/v139/radford21a.html}

\bibitem{radford2019language}
Radford, A., Wu, J., Child, R., Luan, D., Amodei, D., Sutskever, I.: Language
  models are unsupervised multitask learners  (2019)

\bibitem{SaiBhuYanXia2021}
Sain, A., Bhunia, A.K., Yang, Y., Xiang, T., Song, Y.Z.: Stylemeup: Towards
  style-agnostic sketch-based image retrieval. In: Proceedings of the IEEE/CVF
  Conference on Computer Vision and Pattern Recognition. pp. 8504--8513 (2021)

\bibitem{sketchy2016}
Sangkloy, P., Burnell, N., Ham, C., Hays, J.: The {Sketchy} database: Learning
  to retrieve badly drawn bunnies. ACM Transactions on Graphics (proceedings of
  SIGGRAPH)  (2016)

\bibitem{song2017deep}
Song, J., Yu, Q., Song, Y.Z., Xiang, T., Hospedales, T.M.: Deep
  spatial-semantic attention for fine-grained sketch-based image retrieval. In:
  Proceedings of the IEEE international conference on computer vision. pp.
  5551--5560 (2017)

\bibitem{tautkute2019deepstyle}
Tautkute, I., Trzcinski, T., Skorupa, A., Brocki, L., Marasek, K.: Deepstyle:
  Multimodal search engine for fashion and interior design (2019)

\bibitem{tbletchley}
Tiwary, S.: {Turing Bletchley: A Universal Image Language Representation model
  by Microsoft}.
  \url{https://www.microsoft.com/en-us/research/blog/turing-bletchley-a-universal-image-language-representation-model-by-microsoft/}
  (2021), [Online; accessed 7-March-2021]

\bibitem{tursun2021efficient}
Tursun, O., Denman, S., Sridharan, S., Goan, E., Fookes, C.: An efficient
  framework for zero-shot sketch-based image retrieval. arXiv preprint
  arXiv:2102.04016  (2021)

\bibitem{47874}
Vo, N., Jiang, L., Sun, C., Murphy, K., Li, J., Li, F.F., Hays, J.: Composing
  text and image for image retrieval - an empirical odyssey. In: CVPR (2019),
  \url{https://arxiv.org/abs/1812.07119}

\bibitem{WanYanXuHan2017}
Wang, B., Yang, Y., Xu, X., Hanjalic, A., Shen, H.T.: Adversarial cross-modal
  retrieval. In: Proceedings of the 25th ACM international conference on
  Multimedia. pp. 154--162 (2017)

\bibitem{wang2012sketch}
Wang, C., Sun, Z., Zhang, L., Zhang, L.: Sketch2tag: Automatic hand-drawn
  sketch recognition. ACM Conference on Multimedia (January 2012),
  \url{https://www.microsoft.com/en-us/research/publication/sketch2tag-automatic-hand-drawn-sketch-recognition/}

\bibitem{YanJinLinGuo2020}
Yang, Y., Jin, N., Lin, K., Guo, M., Cer, D.: Neural retrieval for question
  answering with cross-attention supervised data augmentation. arXiv preprint
  arXiv:2009.13815  (2020)

\bibitem{yu2016sketch}
Yu, Q., Liu, F., Song, Y.Z., Xiang, T., Hospedales, T.M., Loy, C.C.: Sketch me
  that shoe. In: Proceedings of the IEEE Conference on Computer Vision and
  Pattern Recognition. pp. 799--807 (2016)

\bibitem{zhang2016sketchnet}
Zhang, H., Liu, S., Zhang, C., Ren, W., Wang, R., Cao, X.: Sketchnet: Sketch
  classification with web images. In: Proceedings of the IEEE conference on
  computer vision and pattern recognition. pp. 1105--1113 (2016)

\bibitem{zhang2020context}
Zhang, Q., Lei, Z., Zhang, Z., Li, S.Z.: Context-aware attention network for
  image-text retrieval. In: Proceedings of the IEEE/CVF Conference on Computer
  Vision and Pattern Recognition. pp. 3536--3545 (2020)

\bibitem{ZhaZhaFenZha2020}
Zhang, Z., Zhang, Y., Feng, R., Zhang, T., Fan, W.: Zero-shot sketch-based
  image retrieval via graph convolution network. In: Proceedings of the AAAI
  Conference on Artificial Intelligence. vol.~34, pp. 12943--12950 (2020)

\end{thebibliography}
}

\appendix
\onecolumn

\begin{center}
  {\Large A Sketch Is Worth a Thousand Words:\\[2mm]Image Retrieval with Text
  and Sketch} \\[6mm]
  {\large Supplementary Material}
  \vspace{5mm}
\end{center}


\section{Example Sketches at Various Degrees of Complexity}
\begin{figure}[!h]
    \centering
    \includegraphics[height=0.55\textheight]{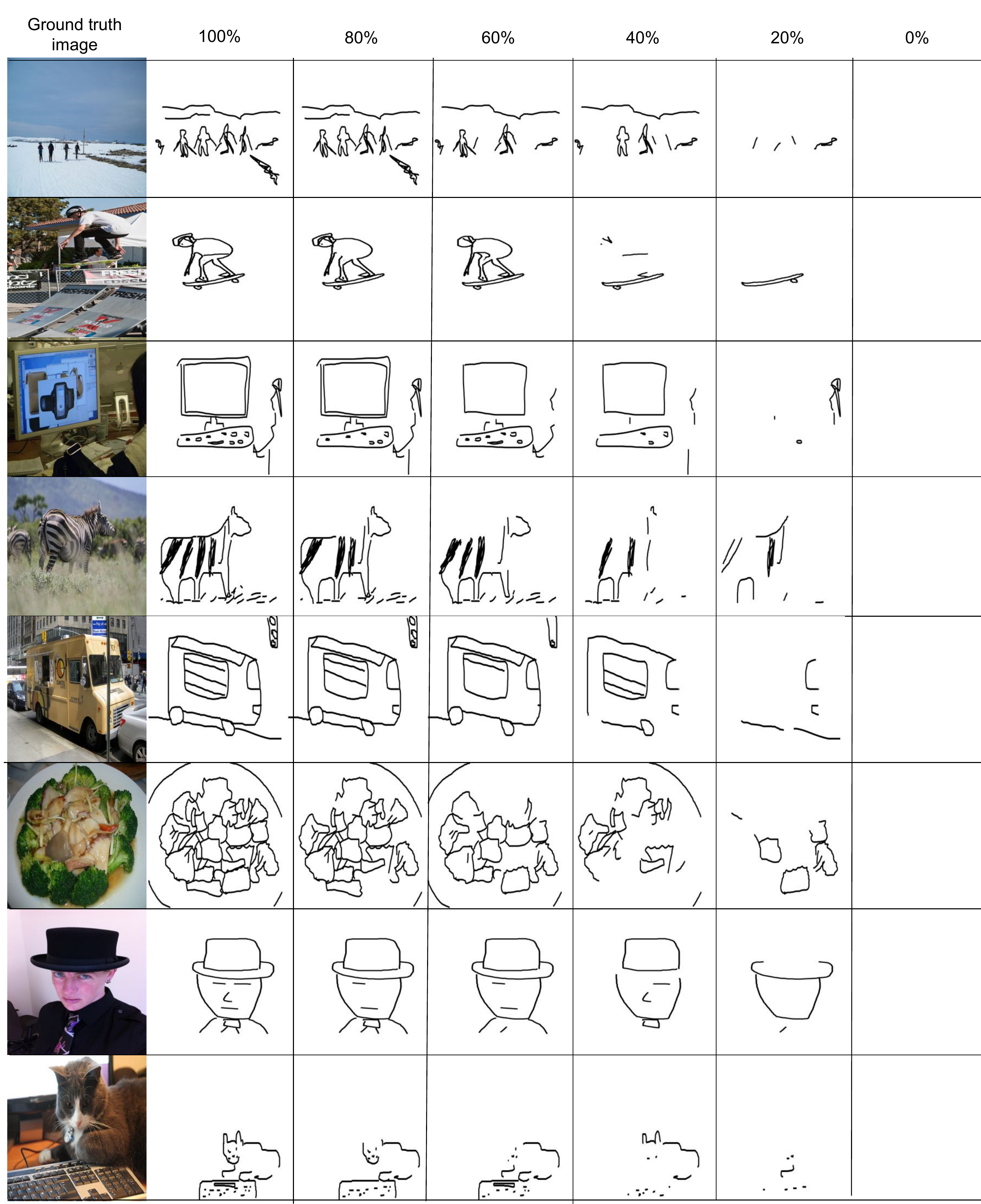}
    \caption{For each collected sketch (100\%), we randomly sub-sample a fraction of strokes to investigate the robustness of our model to incomplete sketches (see \Cref{sec:sketch_complexity}).}
    \label{fig:sketch_complexity_example}
\end{figure}

In this section we show examples of sketches at different levels of complexity, as discussed in \Cref{sec:sketch_complexity}.
For each hand-drawn sketch we collected (100\%), we randomly sub-sample and keep only a fraction of strokes in the sketch.
These incomplete sketches are only used for investigating the robustness of our model to inaccuracy input sketches (see \Cref{sec:sketch_complexity}).
Examples can be found in \Cref{fig:sketch_complexity_example}.

\newpage
\section{Retrieval Results}
\begin{figure}[!h]
    \centering
     \includegraphics[height=0.78\textheight]{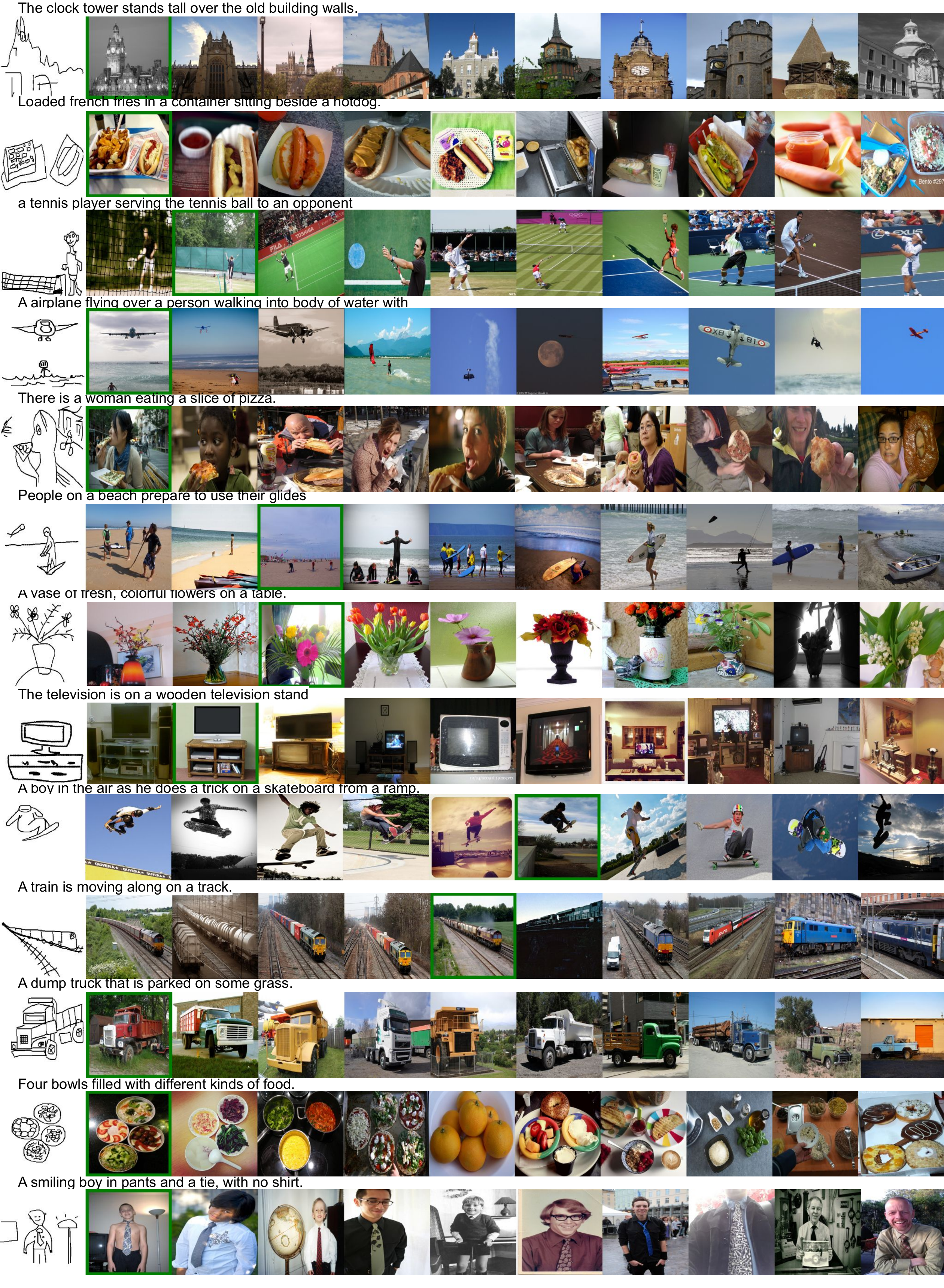}
    \caption{Retrieved images from our model for a number of randomly selected sketch-text pairs.
    }
    \label{fig:qual_res2}
\end{figure}

We present retrieved images from our model for a number of randomly selected input sketch-text pairs in \Cref{fig:qual_res2}.

\newpage
\section{Sketch Captioning with Our Model}

\begin{figure}[!h]
    \centering
    \includegraphics[height=0.52\textheight]{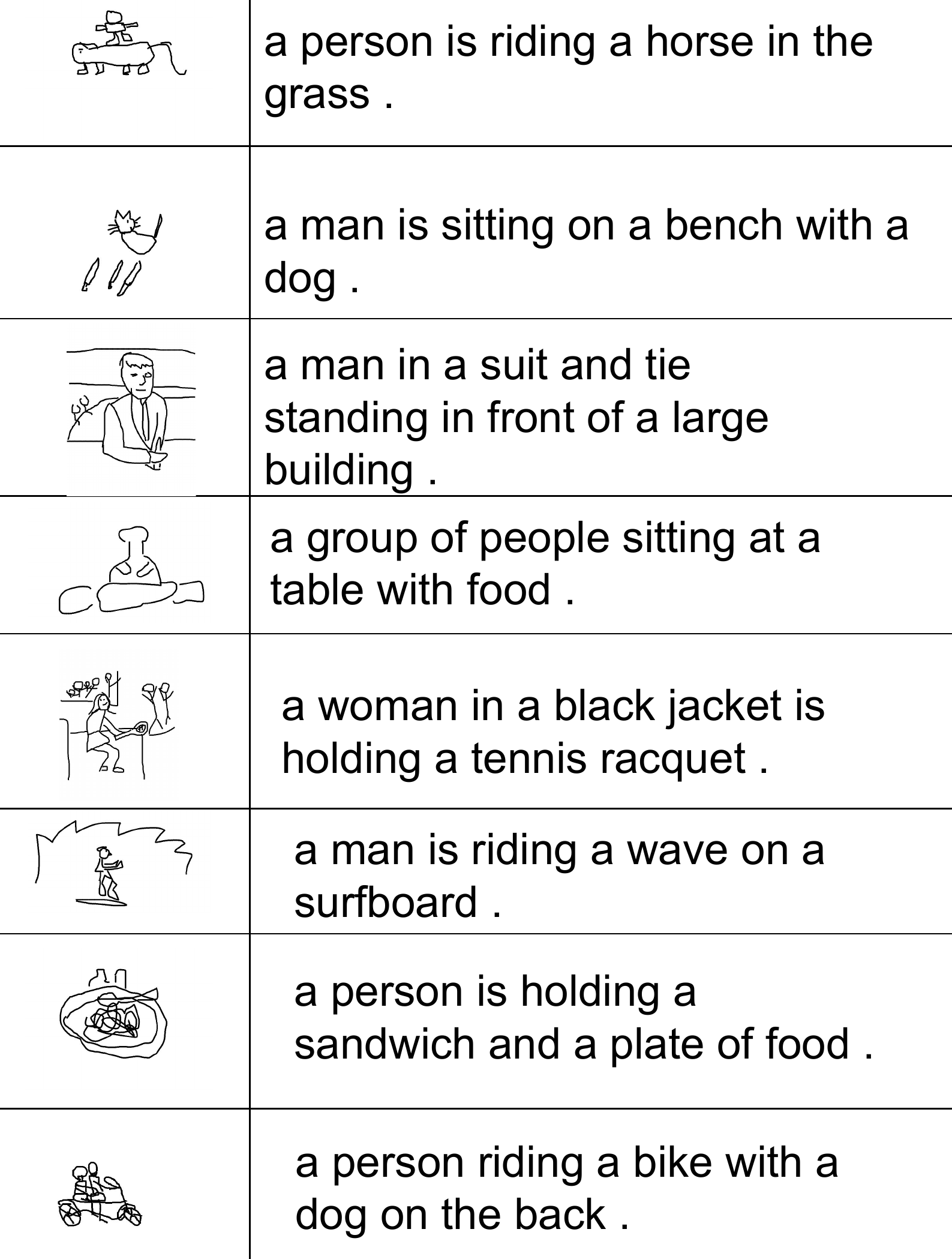}
    \caption{Examples of sketch captions generated from our model.
    The  tendency to start each caption with ``a woman'' or ``a man``  stems from the skewness in the label distribution of the training dataset (COCO) where more than half of images contain people.
    Besides this anticipated issue, our network appears to be able to generate coherent and satisfactorily accurate captions.
    }
    \label{fig:sketch_cap2}
\end{figure}

An interesting application as a result of our caption generation task is caption generation from a sketch.
\Cref{fig:sketch_cap2}  shows examples of these including both success and failure cases.
We observe that the captioning component of our network creates a coherent high-level description overall, but
appears to struggle to produce an accurate description when presented with a sketch of a non-person object.
The network tends to start each caption with ``a man'' or ``a woman`` regardless of the content.
This may be partly due to the skewed  distribution of COCO dataset, where more than half of the training images contain people.
Note that caption generation is not the goal of this work.
The ability to gain insights into what the network sees can be used to design a better training objective for retrieval. This is an interesting topic for future research.

\newpage
\section{Synthetic Sketches}
\begin{figure}[!h]
    \centering
    \includegraphics[width=0.9\textwidth]{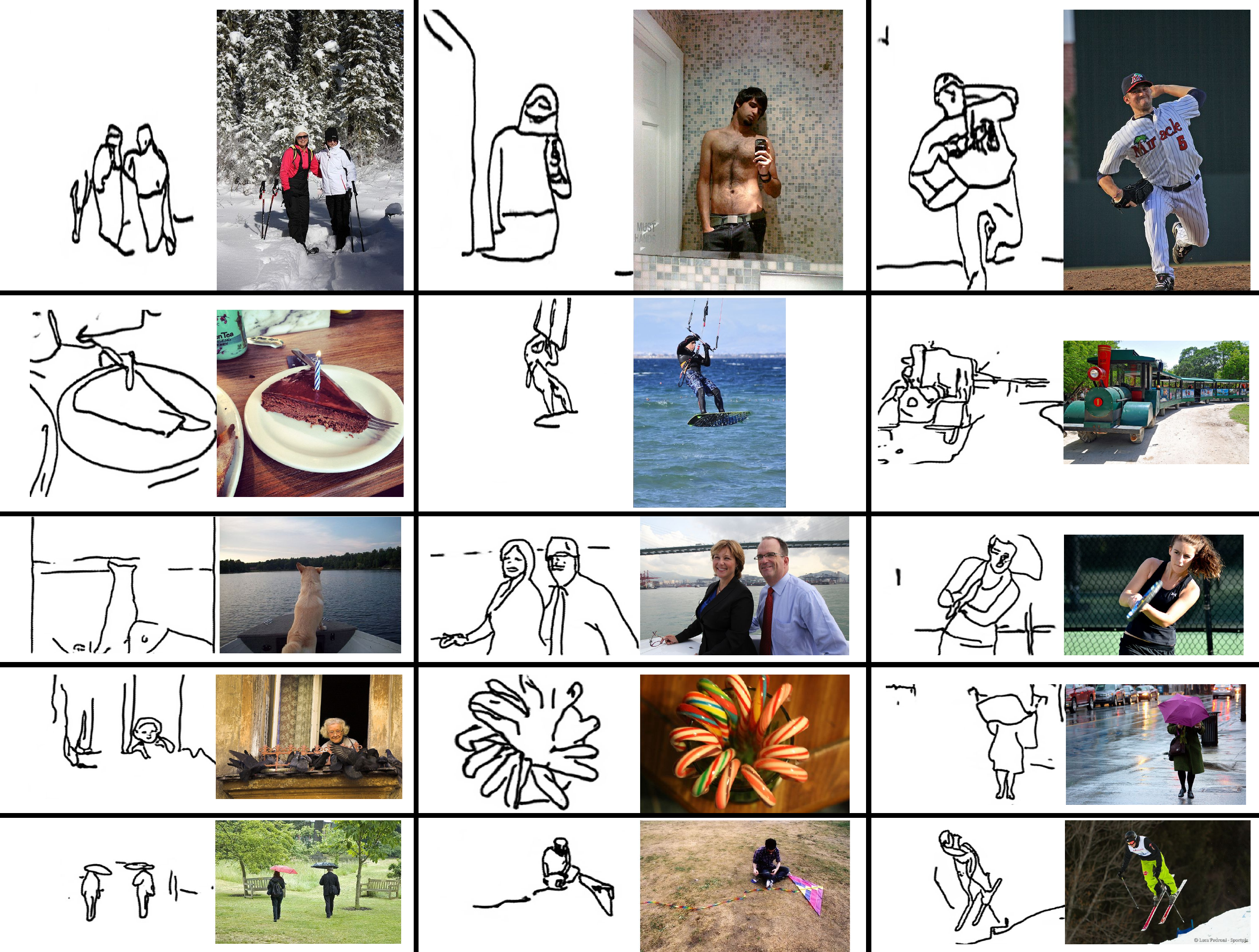}
    \caption{Examples of synthetically generated sketches of \cite{LIPS2019}.
    }
    \label{fig:sketch_syn2}
\end{figure}

Examples of synthetically generated sketches of \cite{LIPS2019} from images in the COCO dataset \cite{coco2018} can be found in \Cref{fig:sketch_syn2}.
Recall from the main text that we use these synthetically sketches to train our model.
For evaluation on COCO, we use hand-drawn sketches.
Details of how we collect hand-drawn sketches can be found in \Cref{sec:sketch_collection}.

\section{Training Details}
For multi-label classification, we first train a multi-label classifier on top of a pre-trained CLIP embedding (freezing all CLIP pre-trained parameters), and use them to initialize the classifier for fine-tuning.
This helps accelerate training as the network is well closer to an optimum than training from scratch with a random initialization.
We do train the decoder for the caption generation task from scratch, however.
We observe that the presence of caption generation as a helper task helps increase  retrieval performance.
We start training of the multi-label classifier with a learning rate of $10^{-4}$ which is then quickly decreased to $10^{-5}$.
In the final network, we use the weight ratio of 10, 1, and 100 for multi label classification, caption generation, and the contrastive learning term respectively.

\newpage
\section{Collecting Hand-Drawn Sketches}
\label{sec:sketch_collection}
While our model is trained with synthetically generated sketches  \cite{LIPS2019}, to investigate the retrieval performance on real data, we collect hand-drawn sketches for images in the COCO test set.
The collected sketches are drawn by Amazon Mechanical Turk (AMT) workers (Turkers).

\begin{figure}[!h]
    \centering
    \includegraphics[width=0.95\textwidth]{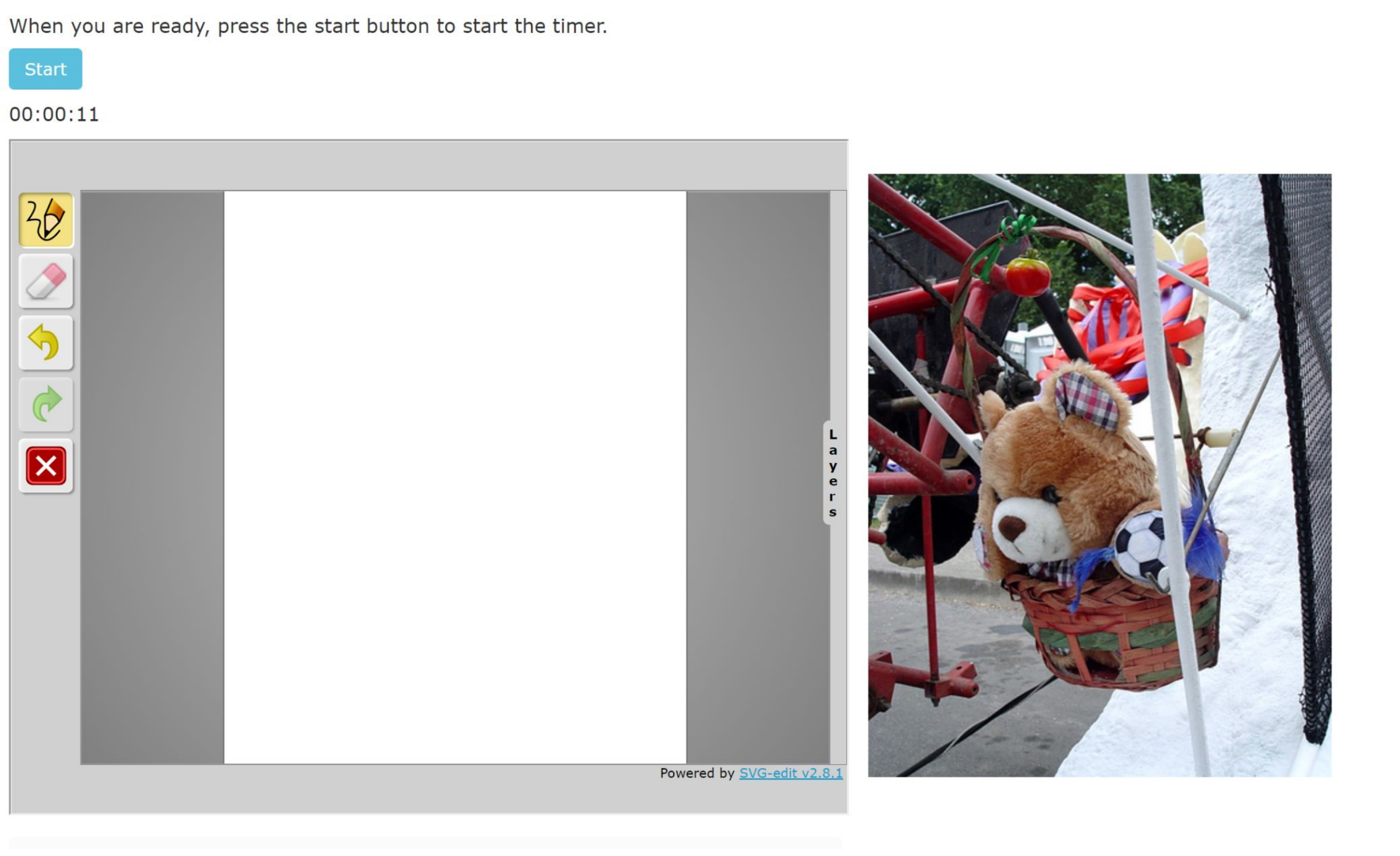}
    \caption{The graphical user interface that workers on Amazon Mechanical Turk use to draw sketches for images in the COCO test set.
    Workers can draw, erase, undo/redo, or clear the entire canvas.
    }
    \label{fig:sketch_amt_gui}
\end{figure}

\begin{figure}[!h]
    \centering
    \includegraphics[width=0.95\textwidth]{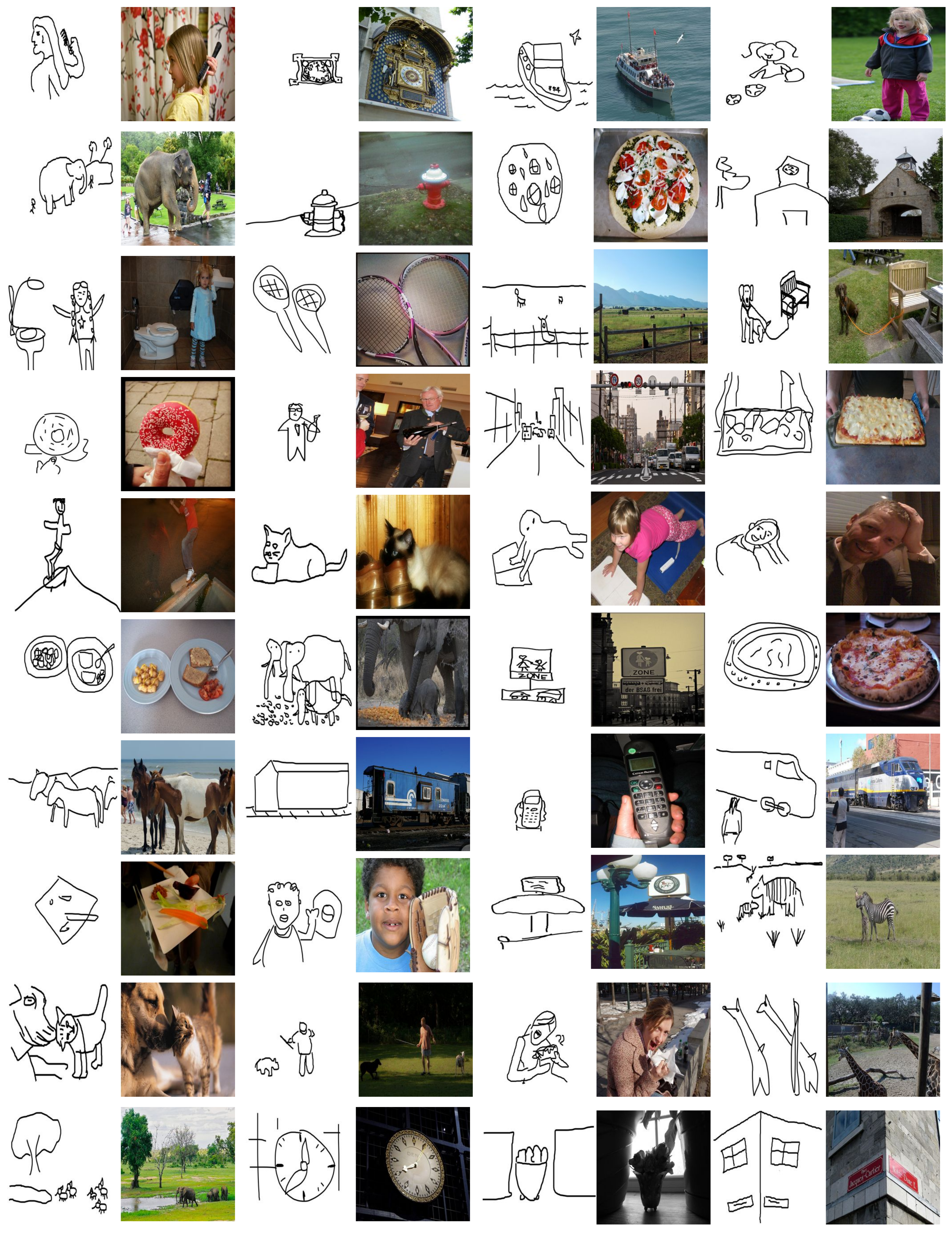}
    \caption{Examples of the hand-drawn sketches collected via Amazon Mechanical Turk.
    }
    \label{fig:sketch_examples}
\end{figure}

The workers draw sketches by going through the following process.
An image from the COCO test set is randomly selected as the target image to draw.
The image is displayed to the worker for 15 seconds and the worker is asked to memorize crucial details in the image.
After 15 seconds, the image disappears and the worker is asked to draw a sketch that represents the image from memory.
Drawing takes place on a graphical user interface (GUI) that we provide (see \Cref{fig:sketch_amt_gui}).
On the provided GUI, the worker can draw, erase, undo a stroke, redo a stroke, or clear the entire canvas.
The size of the sketch pad always matches the size of the target image.
Some example sketches collected can be found in \Cref{fig:sketch_examples}.

\end{document}